\documentclass[10pt,journal,compsoc]{IEEEtran}
\ifCLASSOPTIONcompsoc
  \usepackage[nocompress]{cite}
\else
  \usepackage{cite}
\fi
\usepackage{times}
\usepackage{epsfig}
\usepackage{graphicx}
\usepackage{amsmath}
\usepackage{amssymb}
\usepackage{dsfont}
\usepackage{siunitx}
\usepackage{multirow}
\usepackage{multicol}
\usepackage{subcaption}
\usepackage{adjustbox}
\usepackage{enumitem}
\usepackage{xspace}
\usepackage{url}

\newcommand{\etal}{\textit{et al}. }
\newcommand{\ie}{\textit{i}.\textit{e}. }
\newcommand{\eg}{\textit{e}.\textit{g}.\ }

\newcommand{\figref}[1]{Fig.~\ref{#1}}
\newcommand{\secref}[1]{Section~\ref{#1}}
\newcommand{\tabref}[1]{Table~\ref{#1}}
\graphicspath{{figures/}}

\begin{document}
%
\title{Unsupervised Pre-training for Detection Transformers}
%
%
%
%

\author{Zhigang~Dai,~Bolun~Cai,~Yugeng~Lin,~and~Junying~Chen,~\IEEEmembership{Senior Member,~IEEE}
\IEEEcompsocitemizethanks{\IEEEcompsocthanksitem Z. Dai and J. Chen are with the School of Software Engineering, South China University of Technology, and also with the Key Laboratory of Big Data and Intelligent Robot (SCUT), Ministry of Education, Guangzhou 510006, China.\protect\\
E-mail: zhigangdai@hotmail.com, jychense@scut.edu.cn.
\IEEEcompsocthanksitem B. Cai and Y. Lin are with Tencent Wechat AI, Guangzhou 510630, China.\protect\\
Email: arlencai@tencent.com, lincolnlin@tencent.com.}
\thanks{(Corresponding author: Junying Chen.)}
}
\IEEEtitleabstractindextext{%
\begin{abstract}
DEtection TRansformer (DETR) for object detection reaches competitive performance compared with Faster R-CNN via a transformer encoder-decoder architecture. However, trained with scratch transformers, DETR needs large-scale training data and an extreme long training schedule even on COCO dataset. Inspired by the great success of pre-training transformers in natural language processing, we propose a novel pretext task named random query patch detection in Unsupervised Pre-training DETR (UP-DETR). Specifically, we randomly crop patches from the given image and then feed them as queries to the decoder. The model is pre-trained to detect these query patches from the input image. During the pre-training, we address two critical issues: multi-task learning and multi-query localization. (1) To trade off classification and localization preferences in the pretext task, we find that freezing the CNN backbone is the prerequisite for the success of pre-training transformers. (2) To perform multi-query localization, we develop UP-DETR with multi-query patch detection with attention mask. Besides, UP-DETR also provides a unified perspective for fine-tuning object detection and one-shot detection tasks. In our experiments, UP-DETR significantly boosts the performance of DETR with faster convergence and higher average precision on object detection, one-shot detection and panoptic segmentation. 
Code and pre-training models: \url{https://github.com/dddzg/up-detr}.
\end{abstract}

\begin{IEEEkeywords}
Transformer, Unsupervised Pre-training, Self-supervised Learning, Object Detection, One-shot Detection\end{IEEEkeywords}}

\maketitle

\IEEEdisplaynontitleabstractindextext

%
\IEEEpeerreviewmaketitle

\begin{figure}[t]
\centering
  \includegraphics[width=1.0\linewidth]{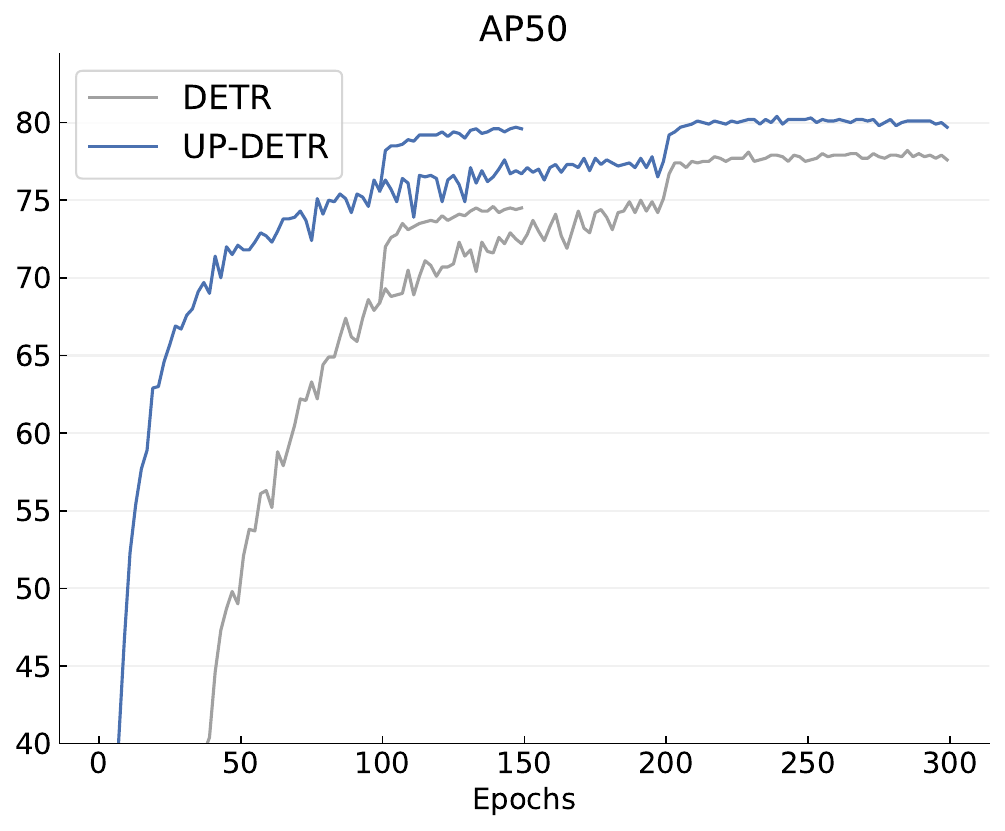}
  \caption{The PASCAL VOC learning curves ($\textrm{AP}_{50}$) of DETR and UP-DETR with the ResNet-50 backbone. Here, they are trained on \texttt{trainval07+12} and evaluated on \texttt{test2007}. We plot the short and long training schedules, and the learning rate is reduced at 100 and 200 epochs respectively for these two schedules. UP-DETR converges much faster with a higher AP than DETR.}
  \label{mAP}
\end{figure}

\IEEEraisesectionheading{\section{Introduction}\label{sec:introduction}}
\IEEEPARstart{D}{Etection} TRansformer (DETR)~\cite{carion2020end} is a recent framework that views object detection as a direct set prediction problem via a transformer encoder-decoder~\cite{vaswani2017attention}. Without hand-designed sample selection~\cite{zhang2020bridging} and non-maximum suppression (NMS), DETR even reaches a competitive performance with Faster R-CNN~\cite{ren2015faster}. However, DETR comes with training and optimization challenges, which need a large-scale training dataset and an extremely long training schedule even on COCO dataset~\cite{lin2014microsoft}. Besides, it is found that DETR performs poorly in PASCAL VOC dataset~\cite{everingham2010pascal} which has insufficient training data and fewer instances than COCO. \figref{mAP} shows the PASCAL VOC learning curves of DETR and our \textbf{Unsupervised Pre-training DETR (UP-DETR)}. UP-DETR converges much faster with a higher AP than DETR.

With well-designed pretext tasks, unsupervised pre-training models achieve remarkable progress in both natural language processing (\eg GPT~\cite{radford2018improving,radford2019language} and BERT~\cite{devlin2018bert}) and computer vision (\eg MoCo~\cite{he2020momentum,chen2020improved} and SwAV~\cite{caron2020unsupervised}). In DETR, the CNN backbone (ResNet-50~\cite{he2016deep} with $\sim$23.2M parameters) has been pre-trained to extract a good visual representation, but the transformer module with $\sim$18.0M parameters has not been pre-trained. 
Therefore, it is a straightforward idea to pre-train the transformer in DETR.

Although unsupervised visual representation learning (\eg contrastive learning) attracts much attention in recent studies~\cite{he2020momentum,chen2020simple,grill2020bootstrap,cao2020parametric,caron2018deep,asano2019self}, existing pretext tasks can not directly apply to pre-train the transformer in DETR. We summarize two main reasons from the following perspectives:

\begin{itemize}
\item[(1)] \textit{\textbf{Different architectures}}: Recently, popular unsupervised learning methods are designed for the backbone (feature extractor) pre-training~\cite{he2020momentum,chen2020simple,grill2020bootstrap} to extract the image feature. However, DETR consists of a backbone and a transformer encoder-decoder. The transformer in DETR is designed to translate the feature into a set of detection targets instead of extracting the visual representation.

\item[(2)] \textit{\textbf{Different feature preference}}: The transformer in DETR mainly focuses on spatial localization learning. Specifically, the transformer is used to translate the instance-level information of the image into coordinates. As for the self-attention in the decoder, it performs a NMS-like mechanism to suppress duplicated bounding boxes. However, existing pretext tasks are designed for 
image instance-based~\cite{he2020momentum,chen2020simple,grill2020bootstrap} or cluster-based~\cite{cao2020parametric,caron2018deep,asano2019self} contrastive learning. These contrastive learning methods mainly focus on feature discrimination rather than spatial localization.
\end{itemize}



As analyzed above, we need to construct a spatial localization related task to pre-train the transformer in DETR. To realize this idea, we propose an Unsupervised Pre-training DETR (UP-DETR) with a novel unsupervised pretext task named \textbf{random query patch detection} to pre-train the detector without any human annotations --- we \textit{randomly} crop multiple \textit{query patches} from the given image, and pre-train the transformer for object \textit{detection} to predict the bounding boxes of these query patches in the given image. During the pre-training procedure, we address two critical issues as follows:


\begin{itemize}
\item[(1)] \textit{\textbf{Multi-task learning}}: Object detection is the coupling of object classification and localization. To avoid query patch detection destroying the classification features, we introduce the \textbf{frozen pre-training backbone} and patch feature reconstruction to preserve the feature discrimination of the transformer. In our experiments, we find that the frozen pre-training backbone is the most important step to preserve the feature discrimination during the pre-training.

\item[(2)] \textit{\textbf{Multi-query localization}}: Different object queries focus on different position areas and box sizes. There is a NMS-like mechanism between different object queries. Considering this property, we propose a \textbf{multi-query patch} detection task with attention mask.
\end{itemize}

Besides, the proposed UP-DETR also provides a unified perspective for object detection and one-shot detection. Just changing the input of the decoder, we can easily fine-tune UP-DETR on object detection and one-shot detection. In our experiments, we pre-train DETR with the random query patch detection task. Our UP-DETR performs much better than DETR using exactly the same architecture on PASCAL VOC~\cite{everingham2010pascal} and COCO~\cite{lin2014microsoft} object detection datasets with faster convergence speed and better average precision. Furthermore, UP-DETR also transfers well with state-of-the-art performance on one-shot detection and panoptic segmentation. 

\section{Related Work}
\subsection{Object Detection}
Positive and negative sample assignment is an important component for object detection frameworks. Two-stage detectors~\cite{ren2015faster,cai2018cascade} and a part of one-stage detectors~\cite{lin2017focal,liu2016ssd} construct positive and negative samples by hand-crafted multi-scale anchors with the IoU threshold and model confidence. Anchor-free one-stage detectors~\cite{tian2019fcos,zhou2019objects,law2018cornernet} assign positive and negative samples to feature maps by a grid of object centers. Zhang \etal demonstrate that the performance gap between them is due to the selection of positive and negative training samples~\cite{zhang2020bridging}. DETR~\cite{carion2020end} is a recent object detection framework that is conceptually simpler without hand-crafted process by direct set prediction~\cite{stewart2016end}, which assigns the positive and negative samples automatically.

Apart from the positive and negative sample selection problem, the trade-off between classification and localization is also intractable for object detection. Zhang \etal illustrate that there is a domain misalignment between classification and localization~\cite{zhang2019towards}. Wu \etal~\cite{wu2020rethinking} and Song \etal~\cite{song2020revisiting} design two head structures for classification and localization. They point out that these two tasks may have opposite feature preferences. As for our pre-training model, it shares feature for classification and localization. Therefore, it is essential to take a well trade-off between these two tasks.

\subsection{Unsupervised Pre-training}
Unsupervised pre-training models always follow two steps: pre-training on a large-scale dataset with the pretext task and fine-tuning the parameters on downstream tasks. For unsupervised pre-training, the pretext task is always invented, and we are interested in the learned intermediate representation rather than the final performance of the pretext task.

To perform unsupervised pre-training, there are various well-designed pretext tasks. For natural language processing, utilizing time sequence relationship between discrete tokens, masked language model~\cite{devlin2018bert}, permutation language model~\cite{yang2019xlnet} and auto regressive model~\cite{radford2018improving,radford2019language} are proposed to pre-train transformers~\cite{vaswani2017attention} for language representation. 
For computer vision, unsupervised pre-training models also achieve remarkable progress recently for visual representation learning, outperforming the supervised learning counterpart in downstream tasks. Instance-based discrimination tasks~\cite{ye2019unsupervised,wu2018unsupervised} and clustering-based tasks~\cite{caron2018deep} are two typical pretext tasks in recent studies. Instance-based discrimination tasks vary mainly on maintaining different sizes of negative samples~\cite{he2020momentum,chen2020simple,grill2020bootstrap} with non-parametric contrastive learning~\cite{hadsell2006dimensionality}. Instance discrimination can also be realized as parametric instance classification~\cite{cao2020parametric}. Moreover, clustering-based tasks vary on offline~\cite{caron2018deep,asano2019self} or online clustering procedures~\cite{caron2020unsupervised}. Our proposed random query patch detection is a novel unsupervised pretext task, which is designed to pre-train the transformer based on the DETR architecture for object detection.

\subsection{Vision Transformer}
Transformers are not only used in detectors, but also applied to backbone designs. The pioneering work is Vision Transformer (ViT)~\cite{Dosovitskiy2021AnII}, which applies a transformer on non-overlapping image patches for image classification. It achieves an speed-accuracy trade-off compared to convolutional neural networks with very large-scale training datasets. Based on this work, there are a lot of works that modify the ViT architecture for better image classification performance~\cite{Yuan2021TokenstoTokenVT,Han2021TransformerIT,Liu2021SwinTH}.

Though, ViT and DETR both use transformers, they are designed for different purposes. ViT uses transformers to capture the relation between image patches for image classification, so there is self-attention between image patches. On the other hand, there are three different attention parts in DETR. The self-attention in the encoder of DETR is similar to ViT, which is designed to capture global relation between image patches. However, the cross attention and self-attention in the decoder of DETR are designed for set prediction of boxes. Therefore, existing pre-training tasks of backbones~\cite{he2020momentum,chen2020improved,he2022masked} can not be simply applied to DETR.

\section{UP-DETR}

\begin{figure*}[ht]
\centering
\begin{subfigure}{0.43\linewidth}
\includegraphics[width=1.0\linewidth]{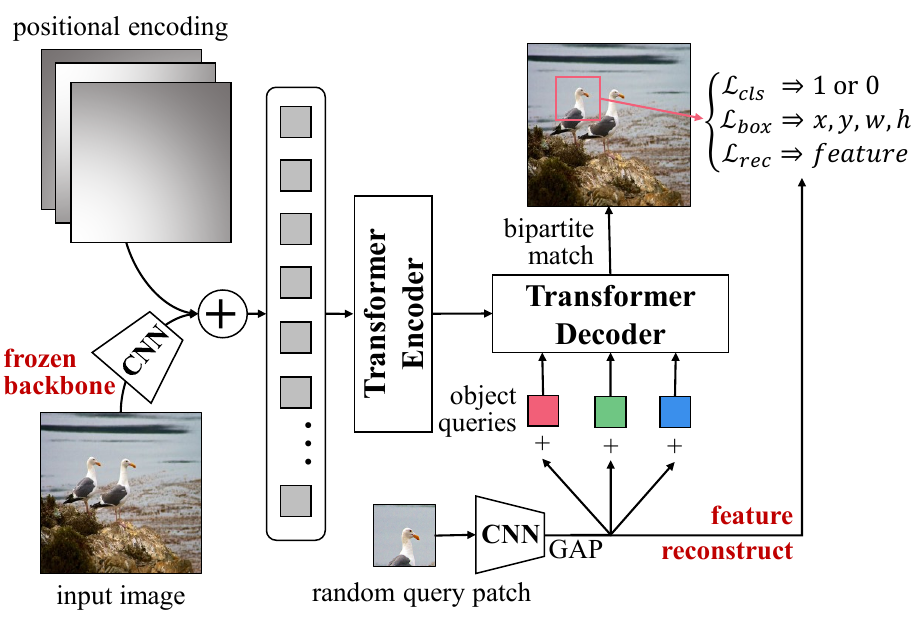} 
\caption{single-query patch (N=3, M=1)}
\label{fig:single}
\end{subfigure}
\begin{subfigure}{0.56\linewidth}
\includegraphics[width=1\linewidth]{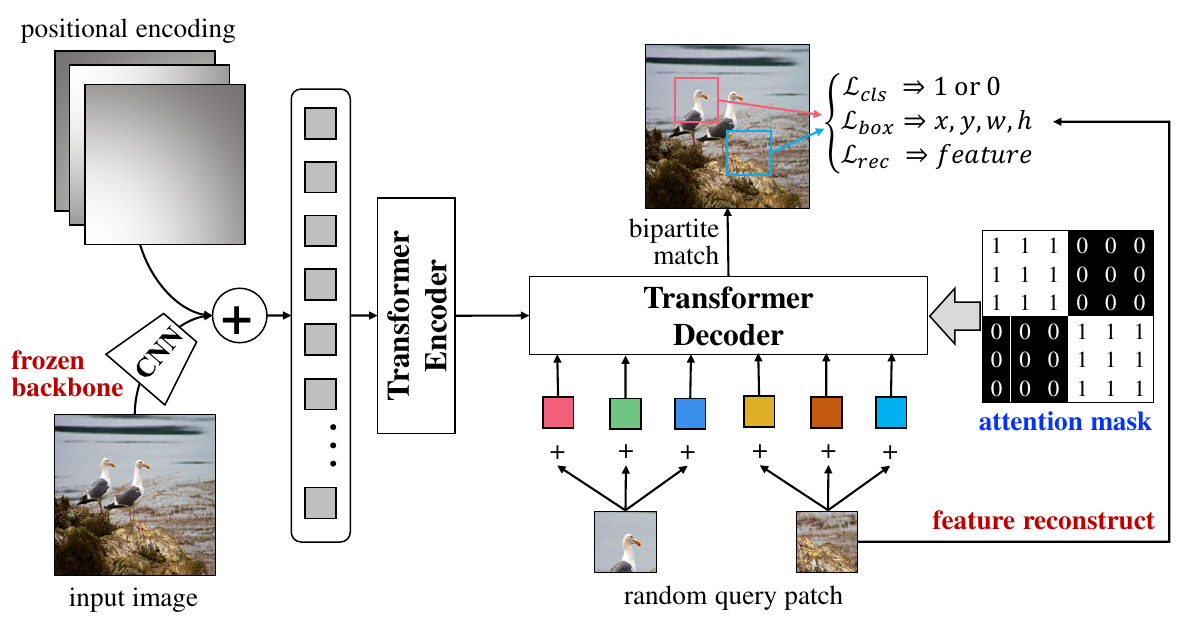}
\caption{multi-query patch (N=6, M=2)}
\label{fig:multi}
\end{subfigure}
  \caption{The pre-training procedure of UP-DETR by random query patch detection. (a) There is only a single-query patch which we add to all object queries. (b) For multi-query patches, we add each query patch to $N/M$ object queries with object query shuffle and attention mask. Note that CNN is not drawn in the decoder of (b) for neatness.} 
  \label{main}
\end{figure*}

The proposed UP-DETR contains pre-training and fine-tuning procedures: (a) the transformer is unsupervisedly \textit{pre-trained} on a large-scale dataset (\eg ImageNet in our experiments) without any human annotations; (b) the entire model is \textit{fine-tuned} with labeled data which is similar as the original DETR~\cite{carion2020end} on the downstream detection related tasks (\eg object detection and one-shot detection). 

\subsection{Pre-training}
In this section, we mainly describe how to pre-train the transformer encoder and decoder with random query patch detection task. 
As shown in \figref{main}, the main idea of random query patch detection is simple but effective. 

\noindent\textbf{Encoder.} Firstly, given an input image, a CNN backbone is used to extract the visual representation with the feature $f \in \mathbb{R}^{C \times H \times W}$, where $C$ is the channel dimension and $H \times W$ is the feature map size.
Here, an image can be treated as a token sequence with the length of $H \times W$ constructed by the CNN extractor. Then, the feature $f$ is added with two-dimensional position encodings and passed to the multi-layer transformer encoder, which is exactly the same as DETR.

\noindent\textbf{Decoder.} Different from DETR, in our pre-training procedure, we randomly crop patches as the query from the input image and record the corresponding coordinates, width and height of the query as the ground truth. 
Therefore, our pre-training process can be done in an unsupervised/self-supervised paradigm. 
Specifically, for the random cropped query patch, the CNN backbone with global average pooling (GAP) extracts the patch feature $p \in \mathbb{R}^{C}$, which is flatten and supplemented with object queries $q \in \mathbb{R}^{C}$ before passing it into a transformer decoder. Finally, the decoder is trained to predict the bounding boxes corresponding to the position of random query patches in the input image. During the pre-training, the feature of query patch is added to multiple object queries, which are fed to the decoder. Note that the \textit{query patch} refers to the cropped patch from the input image but \textit{object query} refers to position embeddings. It can be understood as asking the model to find the query patch through these possible positions (embedded as object queries).
Moreover, the CNN parameters of the input image and query patches are shared in the whole model.

\noindent\textbf{Multi-query Patches.} For fine-tuning object detection tasks, there are multiple object instances in each image (\eg average 7.7 object instances per image in the COCO dataset). In other words, the self-attention module in the decoder transformer learns an NMS-like mechanism during the training. To make a consistency between pre-training and fine-tuning, UP-DETR needs to construct a competition between different object queries with an adjustable box suppression ratio. Fortunately, UP-DETR can be easily extended to multi-query patch detection. Assuming that there are $M$ query patches by random cropping and $N$ object queries, we divide $N$ object queries into $M$ groups, where each query patch is assigned to $N/M$ object queries. In \figref{main}, we illustrate the single-query patch detection and multi-query patch detection. We adopt N=100 and M=10 by default in our pre-training experiments.

\noindent\textbf{Matching.} 
The model infers a prediction with a fixed-set $\hat{y} = \{\hat{y_i}\}^N_{i=1} $ corresponding to $N$ object queries ($N>M$). In other words, we get $N$ pairs of bounding box predictions for these $M$ query patches. For detection tasks, the results are invariant to the permutation of predicted objects. Hence, following DETR~\cite{carion2020end}, we compute the same match cost between the prediction $\hat{y}_{\hat{\sigma}(i)}$ and the ground-truth $y_i$ using Hungarian algorithm~\cite{stewart2016end}, where $\hat{\sigma}(i)$ is the index of $y_i$ computed by the optimal bipartite matching.

\begin{figure*}[ht]
\centering
\begin{subfigure}{0.49\linewidth}
\includegraphics[width=1.0\linewidth]{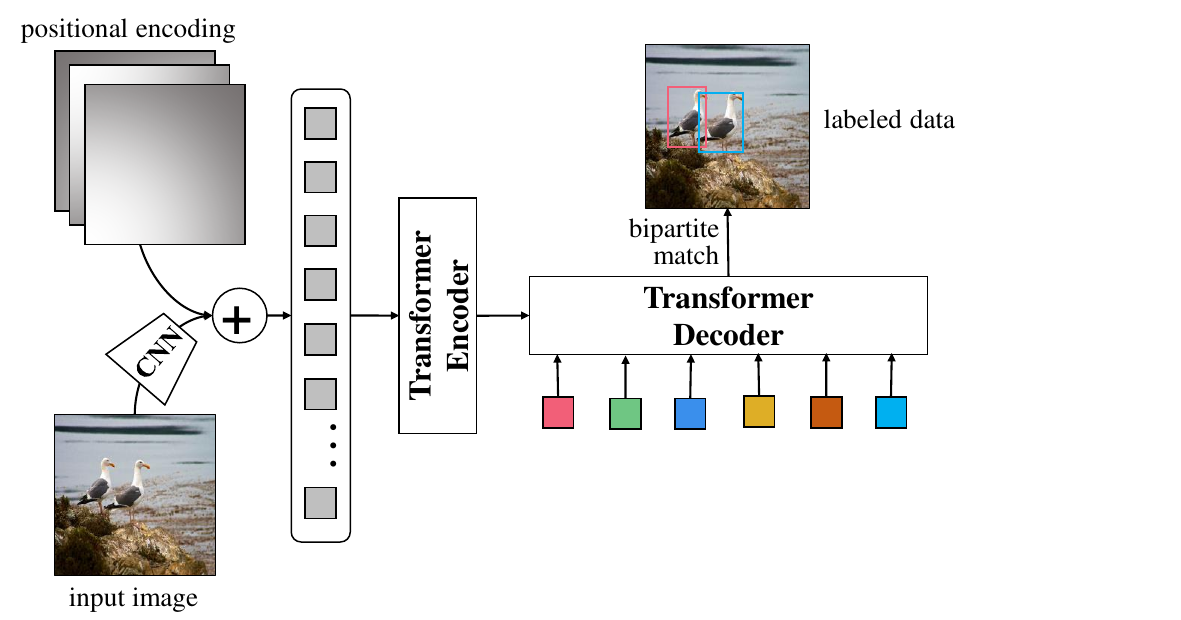} 
\caption{object detection}
\label{fig:obj-det}
\end{subfigure}
\begin{subfigure}{0.49\linewidth}
\includegraphics[width=1\linewidth]{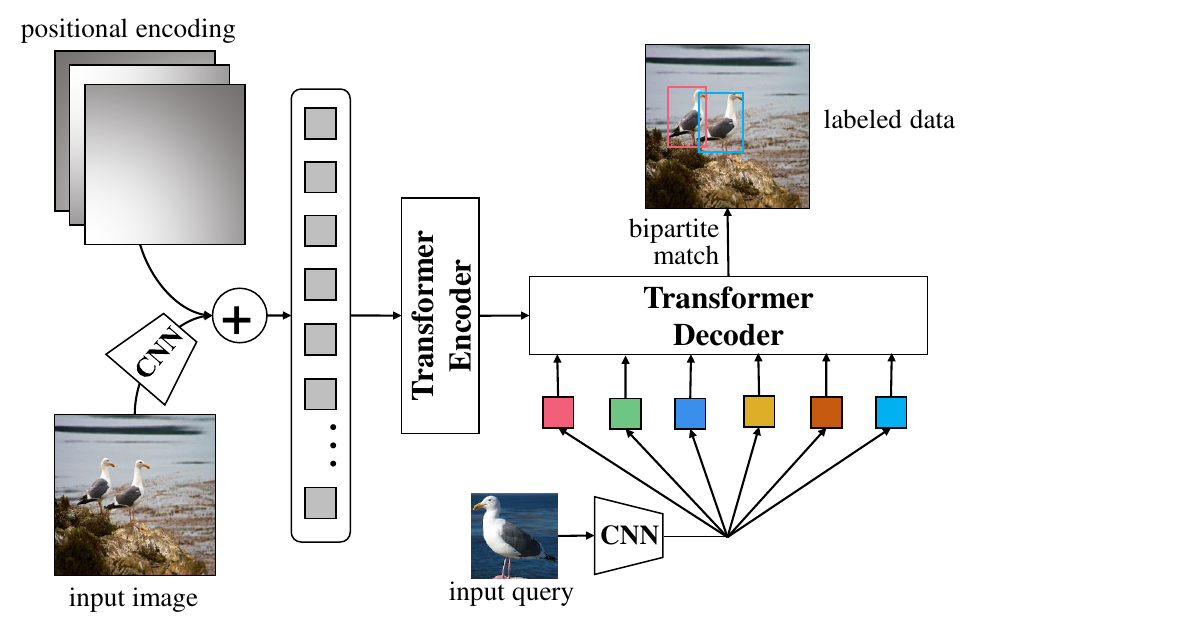}
\caption{one-shot detection}
\label{fig:one-shot-det}
\end{subfigure}
  \caption{The fine-tuning procedure of UP-DETR on object detection and one-shot detection. The only difference is the input of the transformer decoder. (a) Object detection. There are multiple object queries (learnable embeddings) fed into the decoder. (b) One-shot detection. The input query is extracted by the shared CNN, and the feature is added to object queries, which are fed into the decoder.} 
  \label{fine-tune-proc}
\end{figure*}

\noindent\textbf{Loss.} For the loss calculation, the predicted result $\hat{y}_i = (\hat{c}_i \in \mathbb{R}^2 ,\hat{b}_i \in \mathbb{R}^4 ,\hat{p}_i \in \mathbb{R}^{C} )$ consists of three elements: $\hat{c}_i$ is the binary classification of matching the query patch ($c_i=1$) or not ($c_i=0$) for each object query; $\hat{b}_i$ is the vector that defines the box center coordinates and its width and height as $\{x,y,w,h\}$, which are re-scaled relative to the image size; and $\hat{p}_i$ is the reconstructed feature with $C=2048$ for the ResNet-50 backbone typically. With the above definitions, the Hungarian loss for all matched pairs is defined as:
\begin{align}
\mathcal{L}(y, \hat{y}) =& \sum_{i=1}^{N}[\lambda_{\{c_i\}}\mathcal{L}_{cls}(c_i,\hat{c}_{\hat{\sigma}(i)}) + \mathds{1}_{\{c_{i} =1\}} \mathcal{L}_{box}(b_i,\hat{b}_{\hat{\sigma}(i)}) \nonumber\\
 & +\mathds{1}_{\{c_{i} =1\}} \mathcal{L}_{rec}(p_i,\hat{p}_{\hat{\sigma}(i)})].
\end{align}
Here, $\mathcal{L}_{cls}$ is the cross entropy loss over two classes (match the query patch $vs.$ not match), and the class balance weight $\lambda_{\{c_i=1\}}=1$ and $\lambda_{\{c_i=0\}}=M/N$. $\mathcal{L}_{box}$ is a linear combination of $\ell_{1}$ loss and the generalized IoU loss with the same weight hyper-parameters as DETR~\cite{carion2020end}. $\mathcal{L}_{rec}$ is the reconstruction loss proposed in this paper and only used during the pre-training. It is designed to 
balance the feature preference of classification and localization during the pre-training, which will be further discussed in ~\secref{rud}.

\subsubsection{Frozen Pre-training Backbone}
In our experiments, we find that the CNN backbone seriously affects the transformer pre-training. In other words, the model can not converge well and performs bad if we pre-train the CNN backbone and the transformer from scratch together with random query patch detection. This problem also appears in the original DETR\footnote{https://github.com/facebookresearch/detr/issues/157}.

Furthermore, object detection is the coupling of object classification and localization, where these two tasks always have different feature preferences~\cite{zhang2019towards,wu2020rethinking,song2020revisiting}. However, our pre-training task only focuses on localization instead of feature classification. So, we freeze the pre-training CNN backbone during the transformer pre-training in our experiments. Stable backbone parameters are beneficial to the transformer pre-training, and accelerate the model pre-training process. In \secref{freeze_CNN}, we will analyze and verify the necessity of them with experiments. We argue that the frozen pre-training backbone is the most essential step to the success of UP-DETR.



\subsubsection{Tricks}
\label{rud}
In our preliminary experiments, we tried some tricks with our prior knowledge in terms of random query patch detection. Here, we discuss two tricks during the pre-training, which are reasonable but not critical to the success of UP-DETR.

\noindent\textbf{Patch Feature Reconstruction.}
Our pre-training task only focuses on localization instead of feature classification. In other words, missing an explicit branch for the classification task.
So, with the frozen pre-training backbone, we propose a feature reconstruction loss term $\mathcal{L}_{rec}$ to preserve classification feature during localization pre-training. The motivation of this loss term is to preserve the feature discrimination extract by CNN after passing feature to the transformer. $\mathcal{L}_{rec}$ is the mean squared error between the $\ell_{2}$-normalized patch feature extracted by the CNN backbone, which is defined as follows:
\begin{align}
\mathcal{L}_{rec}(p_i,\hat{p}_{\hat{\sigma}(i)}) &= \left\| \frac{p_i} {\left\| p_i \right\|_{2}}- \frac{\hat{p}_{\hat{\sigma}(i)}}{\left\| \hat{p}_{\hat{\sigma}(i)} \right\|_{2}} \right\|_{2}^{2}.
\end{align}
With the frozen CNN backbone, patch feature reconstruction slightly improves the fine-tuning performance as shown in \secref{freeze_CNN}.

\noindent\textbf{Attention Mask.} All the query patches are randomly cropped from the image. Therefore, they are independent without any relations. For example, the bounding box regression of the first cropping is not concerned with the second cropping. To satisfy the independence of query patches, we utilize an attention mask matrix to control the interactions between different object queries. The mask matrix $\mathbf{X} \in \mathbb{R}^{N \times N}$ is added to the softmax layer of self-attention~\cite{vaswani2017attention} in the decoder as $softmax\left(\mathbf{Q}\mathbf{K}^{\top}/{\sqrt{d_{k}}}+\mathbf{X}\right) \mathbf{V}$, where $\mathbf{Q}=\mathbf{K}=\mathbf{V}$ and they refer to the same set of the object query representation in the decoder. Similar to the token mask in UniLM \cite{dong2019unified}, the attention mask is defined as:
\begin{equation}\label{eq:mask}
X_{i, j}=\left\{\begin{array}{ll}
0, & \text {\it{i}, \it{j} in the same group} \\
-\infty, & \text {otherwise}
\end{array}\right.,
\end{equation}
where $X_{i, j}$ determines whether the object query $q_i$ participates the interaction with the object query $q_j$. For intuitive understanding, the attention mask in \figref{fig:multi} displays $1$ and $0$ corresponding to $0$ and $-\infty$ in \eqref{eq:mask}, respectively. Attention mask also slightly leads to a lower loss as demonstrated in \secref{attmask}.

\subsection{Fine-tuning}
UP-DETR provides a unified perspective in terms of detection related tasks. It can be easily fine-tuned into object detection or one-shot detection by changing the model input. \figref{fine-tune-proc} shows the fine-tuning procedure on object detection and one-shot detection. The CNN, transformer encoder-decoder and object queries are exactly the same for object detection and one-shot detection tasks, which are initialized from the pre-training procedure. The whole model weights are fine-tuned with the labeled data. The only difference in architecture between these two down-stream tasks is the input of transformer decoder. Noting that feature reconstruction loss is never used in the fine-tuning procedure.

\noindent\textbf{Object detection.} Given an input image, the model is required to predict to the set of objects with the bounding boxes and the corresponding categories. Therefore, the fine-tuning procedure of UP-DETR is exactly the same with the training procedure of DETR. There are multiple object queries (learnable embeddings) used as the input fed into the decoder. Here, different object queries learn different spatial specialization and area preference. Benefiting from the design of object query, the model can predict multiple objects in parallel. With supervised object detection data, we compute the match cost between model prediction and the annotated ground-truth. 

\noindent\textbf{One-shot detection.} Given an input image and a query image, the model is required to predict the objects with bounding boxes. The objects should be semantically similar to the query image. Typically, the query image is constructed by patch from different image with the same category. For one-shot detection, the query image is extracted by the CNN (shared with the input image), and the patch feature is added to all object queries. In one-shot detection, we only care about the matching result (\ie match to the query or not) of the bounding boxes instead of the specific category in the object detection. So, in the loss and match cost computation, there is a binary classification output in one-shot detection instead of multiple category classification in object detection.

\section{Experiments}
We pre-train the model using ImageNet~\cite{deng2009imagenet} and fine-tune the parameters on PASCAL VOC~\cite{everingham2010pascal} and COCO~\cite{lin2014microsoft}. In all experiments, we implement the UP-DETR model (41.3M parameters) with ResNet-50 backbone, 6 transformer encoder, 6 decoder layers of width 256 with 8 attention heads. Referring to the open source code of DETR\footnote{https://github.com/facebookresearch/detr}, we use the same hyper-parameters in the proposed UP-DETR and our DETR re-implementation. We annotate R50 and R101 short for ResNet-50 and ResNet-101. Note that UP-DETR and DETR have exactly the same model architecture, match cost and loss calculation. So, they have exactly the same FLOPs, parameters and running FPS.

\noindent\textbf{Pre-training setup.} UP-DETR is unsupervisedly pre-trained on the 1.28M ImageNet training set without any ImageNet labels. The CNN backbone (ResNet-50) is also unsupervisedly pre-trained with SwAV strategy~\cite{caron2020unsupervised}, and its parameters are frozen during the transformer pre-training. As the input image from ImageNet is relatively small, we resize it such that the shortest side is within $[320, 480]$ pixels while the longest side is at most 600 pixels. 
For the given image, we crop the query patches with random coordinate, height and width, which are resized to $128\times128$ pixels and transformed with SimCLR style~\cite{chen2020simple} without horizontal flipping, including random color distortion and Gaussian blurring. In the early stage of our work, we conducted experiments about different sampling strategies, \textit{i.e.}, grid sampling, and random sampling with different hyper-parameters. However, we noticed that there were no significant differences between different sampling strategies. Therefore, we use random sampling to illustrate the generalization capability of our method.
Moreover, AdamW~\cite{loshchilov2017decoupled} is used to optimize the UP-DETR, with the initial learning rate of \num{1e-4} and the weight decay of \num{1e-4}. We use a mini-batch size of 256 on eight Nvidia V100 GPUs to train the model for 60 epochs with the learning rate multiplied by 0.1 at 40 epochs.

\noindent\textbf{Fine-tuning setup.} The model is initialized with the pre-trained UP-DETR parameters and fine-tuned for all the parameters (including CNN) on PASCAL VOC and COCO. We fine-tune the model with the initial learning rate of \num{1e-4} for the transformer and \num{5e-5} for the CNN backbone, and the other settings are the same as DETR~\cite{carion2020end} on eight V100 GPUs with four images per GPU. The model is fine-tuned with short/long schedule for 150/300 epochs and the learning rate is multiplied by 0.1 at 100/200 epochs, respectively.

\noindent\textbf{Gap between pre-training and fine-tuning.} Object-level pre-training can boost the fine-tuning performance in object detection task~\cite{bar2022detreg}, but such method usually requires object prior knowledge, \textit{e.g.}, using a selective search algorithm to get object proposals during pre-training. In our work, we pre-train the model on ImageNet dataset at patch level, as this is a common practice in pre-training on ImageNet and transferring to COCO or VOC datasets\cite{he2020momentum,he2022masked}. However, the gap between pre-training and fine-tuning always exists. For example, ImageNet images mostly contain one object, but COCO images always contain multiple objects. The image size of ImageNet images is relatively small, but the image size of COCO images is large. Moreover, the pre-training task is conducted at patch level with random sampling in this work, but the fine-tuning task is supervised on labeled data. Although the aforementioned gap between pre-training and fine-tuning cannot be eliminated, our method has good generalization capability using random sampling strategy without any prior knowledge.

\subsection{PASCAL VOC Object Detection}
\label{voc}
\noindent\textbf{Setup.} The model is fine-tuned on PASCAL VOC \texttt{trainval07+12} ($\sim$16.5k images) and evaluated on \texttt{test2007}. We report COCO-style metrics, including AP, $\textrm{AP}_{50}$ (default VOC metric) and $\textrm{AP}_{75}$. For a full comparison, we also report the result of Faster R-CNN with the R50-C4 backbone~\cite{caron2020unsupervised}, which performs much better than R50 (C5 stage)~\cite{li2017fully}. DETR with R50-C4 significantly increases the computational cost as compared with R50, so we fine-tune UP-DETR with R50 backbone. Even though, UP-DETR still performs well. To emphasize the effectiveness of pre-training models, we report the results of 150 and 300 epochs for both DETR (from the random initialized transformer) and UP-DETR (from the pre-trained transformer).

\begin{table}[!t]
\caption{Object detection results trained on PASCAL VOC \texttt{trainval07+12} and evaluated on \texttt{test2007}. DETR and UP-DETR use R50 backbone and Faster R-CNN uses R50-C4 backbone. The values in the brackets are the gaps compared to DETR with the same training schedule.}
\centering
\label{voc_exp}
\begin{tabular}{l|lll}
\hline\hline
Model/Epoch & AP & $\textrm{AP}_{50}$ & $\textrm{AP}_{75}$ \\
\hline
Faster R-CNN & 56.1 & \textbf{82.6} & \textbf{62.7} \\
\hline
DETR/150      & 49.9 & 74.5 & 53.1 \\
UP-DETR/150     & \textbf{56.1} {\small (+6.2)}& 79.7 {\small(+5.2)}& 60.6 {\small (+7.5)}\\
\hline
DETR/300    & 54.1 & 78.0 & 58.3 \\
UP-DETR/300     & \textbf{57.2} {\small (+3.1)}& 80.1 {\small(+2.1)}& 62.0 {\small(+3.7)}\\
\hline\hline
\end{tabular}
\end{table}

\begin{figure*}[ht]
\centering
\begin{subfigure}{0.48\linewidth}
\includegraphics[width=0.96\linewidth]{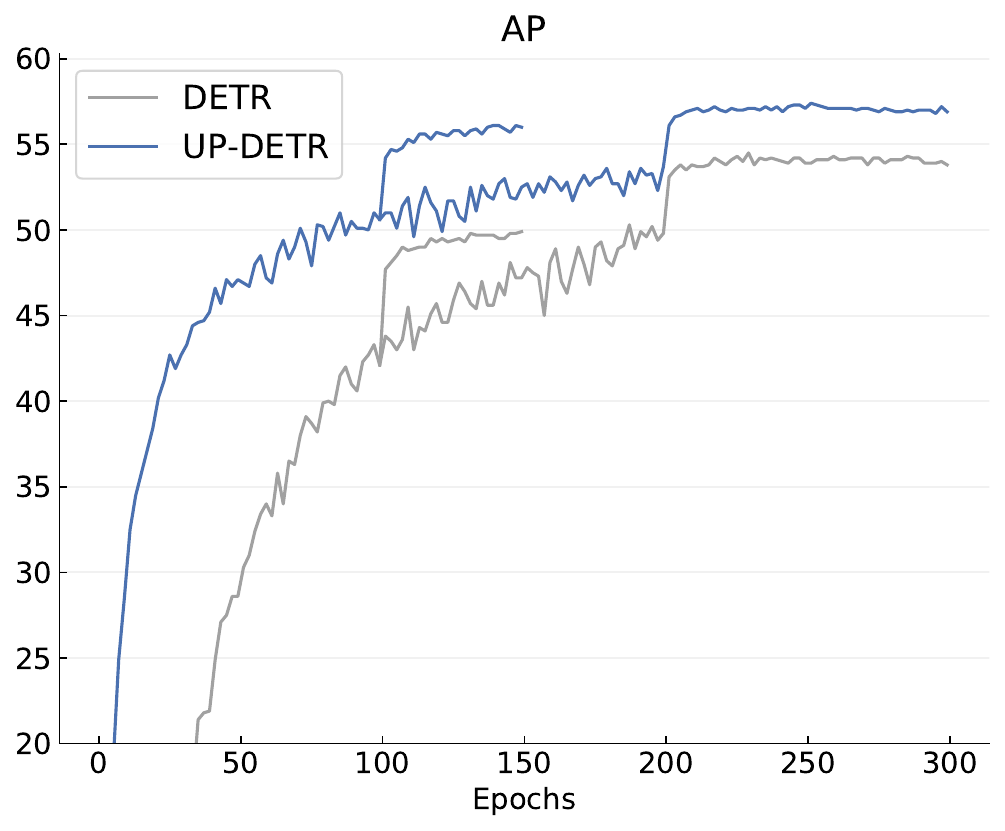} 
\caption{AP learning curves of PASCAL VOC}
\label{fig:vocAP}
\end{subfigure}
\begin{subfigure}{0.48\linewidth}
\includegraphics[width=0.96\linewidth]{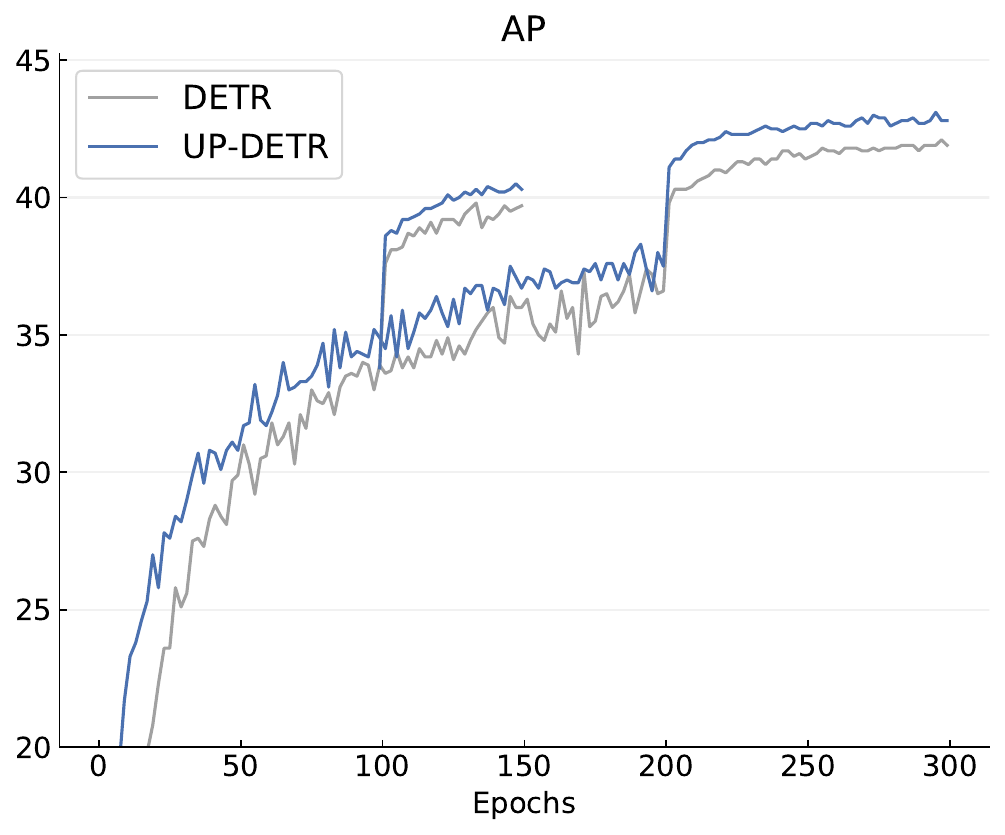}
\caption{AP learning curves of COCO}
\label{fig:cocoAP}
\end{subfigure}
\caption{AP (COCO style) learning curves of DETR and UP-DETR on PASCAL VOC and COCO datasets. Models are trained with the SwAV pre-training ResNet-50 for 150 and 300 epochs, and the learning rate is reduced at 100 and 200 epochs, respectively.}
\label{fig:APcurves}
\end{figure*}

\begin{table*}[!h]
\caption{Object detection results trained on COCO \texttt{train2017} and evaluated on \texttt{val2017}. Faster R-CNN, DETR and UP-DETR are performed under comparable settings. $\dagger$ for values evaluated on COCO \texttt{test-dev}, which are always slightly higher than \texttt{val2017}. The values in the brackets are the gaps compared to DETR (SwAV CNN) with the same training schedule.}
\label{coco}
\centering
\begin{tabular}{lll|llllll}
\hline\hline
Model & Backbone & Epochs & AP   & $\textrm{AP}_{50}$ & $\textrm{AP}_{75}$ & $\textrm{AP}_{S}$& $\textrm{AP}_{M}$& $\textrm{AP}_{L}$ \\\hline
Faster R-CNN $\dagger$~\cite{lin2017feature} & R101-FPN & -& 36.2 & 59.1 & 39.0 & 18.2 & 39.0 & 48.2 \\
Mask R-CNN $\dagger$~\cite{he2017mask} & R101-FPN & -& 38.2 & 60.3 & 41.7 & 20.1 & 41.1 & 50.2 \\
Grid R-CNN $\dagger$~\cite{lu2019grid} & R101-FPN & -& 41.5 & 60.9 & 44.5 & 23.3 & 44.9 & 53.1 \\
Double-head R-CNN~\cite{wu2020rethinking} & R101-FPN & -& 41.9 & 62.4 & 45.9 & 23.9 & 45.2 & 55.8 \\
RetinaNet $\dagger$~\cite{lin2017focal} & R101-FPN & -& 39.1 & 59.1 & 42.3 & 21.8 & 42.7 & 50.2 \\
FCOS $\dagger$~\cite{tian2019fcos} &  R101-FPN & -& 41.5 & 60.7 & 45.0 & 24.4 & 44.8 & 51.6 \\
DETR~\cite{carion2020end} &  R50 & 500 & 42.0 & 62.4 & 44.2 & 20.5 & 45.8 & 61.1 \\
\hline
\hline
Faster R-CNN & R50-FPN & $3\times$& 40.2 & \textbf{61.0} & \textbf{43.8} & \textbf{24.2} & 43.5 &  52.0 \\
DETR (Supervised CNN) & R50 & 150 & 39.5 & 60.3 & 41.4 & 17.5 & 43.0 & 59.1 \\
DETR (SwAV CNN)~\cite{caron2020unsupervised} & R50 & 150 & 39.7 & 60.3 & 41.7 & 18.5 & 43.8 & 57.5 \\
\textbf{UP-DETR} & R50 & 150 & \textbf{40.5} \small (+0.8) & 60.8 & 42.6  & 19.0 & \textbf{44.4} & \textbf{60.0} \\\hline
\hline
Faster R-CNN & R50-FPN & $9\times$ & 42.0 & 62.1 & 45.5 & \textbf{26.6} & 45.4 & 53.4 \\
DETR (Supervised CNN) & R50 & 300 & 40.8 & 61.2 & 42.9 & 20.1 & 44.5 & 60.3 \\
DETR (SwAV CNN)~\cite{caron2020unsupervised} & R50 & 300 & 42.1 & 63.1 & 44.5 & 19.7 & 46.3 & 60.9 \\
\textbf{UP-DETR} & R50 & 300 & \textbf{43.1} \small (+1.0) & \textbf{63.4} & \textbf{46.0} & 21.6 & \textbf{46.8} & \textbf{62.4} \\
\hline\hline
\end{tabular}
\end{table*}

\noindent\textbf{Results.} \tabref{voc_exp} shows the object detection results on PASCAL VOC dataset. 
We find that DETR performs poorly on PASCAL VOC, which is much worse than Faster R-CNN by a large gap in all metrics. Due to the relatively small-scale data in VOC, the pre-training transformer of UP-DETR significantly boosts the performance of DETR for both short and long schedules: up to \textbf{+6.2 (+3.1)} AP, \textbf{+5.2 (+2.1)} $\textrm{AP}_{50}$ and \textbf{+7.5 (+3.7)} $\textrm{AP}_{75}$ for 150 (300) epochs, respectively. Moreover, UP-DETR (R50) achieves a comparable result to Faster R-CNN (R50-C4) with better AP. We find that both UP-DETR and DETR perform a little worse than Faster R-CNN in $\textrm{AP}_{50}$ and $\textrm{AP}_{75}$. This may come from different ratios of feature maps (C4 for Faster R-CNN) and no NMS post-processing (NMS lowers AP but slightly improves $\textrm{AP}_{50}$).

\figref{fig:vocAP} shows the AP (COCO style) learning curves on VOC. UP-DETR significantly speeds up the model convergence. After the learning rate reduced, UP-DETR significantly boosts the performance of DETR with a large AP improvement. Noting that UP-DETR obtains 56.1 AP after 150 epochs, however, its counterpart DETR (scratch transformer) only obtains 54.1 AP even after 300 epochs and does not catch up even training longer. It suggests that pre-training transformer is indispensable on insufficient training data (\ie $\sim 16.5$K images on VOC).

\subsection{COCO Object Detection}
\noindent\textbf{Setup.} The model is fine-tuned on COCO \texttt{train2017} ($\sim$118k images) and evaluated on \texttt{val2017}. There are lots of small objects in COCO dataset, where DETR performs poorly~\cite{carion2020end}. Therefore, we report AP, $\textrm{AP}_{50}$, $\textrm{AP}_{75}$, $\textrm{AP}_{S}$, $\textrm{AP}_{M}$ and $\textrm{AP}_{L}$ for a comprehensive comparison. Moreover, we also report the results of highly optimized Faster R-CNN with feature pyramid network (FPN) with short (3$\times$) and long (9$\times$) training schedules, which are known to improve the performance results~\cite{he2019rethinking}. To avoid supervised CNN bringing supplementary information, we use SwAV pre-training CNN as the backbone of UP-DETR without any human annotations. 

\noindent\textbf{Results.} 
\tabref{coco} shows the object detection results on COCO dataset. 
With 150 epoch schedule, UP-DETR outperforms DETR (SwAV CNN) by 0.8 AP and achieves a comparable performance as compared with Faster R-CNN (R50-FPN) (3$\times$ schedule). With 300 epoch schedule, UP-DETR obtains \textbf{43.1} AP on COCO, which is 1.0 AP better than DETR (SwAV CNN) and 1.1 AP better than Faster R-CNN (R50-FPN) (9$\times$ schedule). Overall, UP-DETR comprehensively outperforms DETR in detection of small, medium and large objects with both short and long training schedules. Regrettably, UP-DETR is still slightly lagging behind Faster R-CNN in $\textrm{AP}_{S}$, because of the lacking of FPN-like architecture~\cite{lin2017feature}.

\begin{table*}[]
\caption{One-shot detection results trained on VOC \texttt{2007train val} and \texttt{2012train val} sets and evaluated on VOC \texttt{2007test} set.}
\label{OSD}
\vspace{-7pt}
\begin{center}
\begin{adjustbox}{max width=1\linewidth}
\setlength{\tabcolsep}{0.8mm}{
\begin{tabular}{c|c|c|c|c|c|c|c|c|c|c|c|c|c|c|c|c|c|c|c|c|c|c}
\hline\hline
\multirow{2}{*}{Model}  & \multicolumn{17}{c|}{seen class} & \multicolumn{5}{c}{unseen class} \\ \cline{2-23}
&plant& sofa& tv& car& bottle& boat& chair& person& bus& train& horse& bike& dog& bird& mbike& table& \textbf{AP$^{50}$}& cow& sheep& cat& aero& \textbf{AP$^{50}$} \\
\hline
SiamFC~\cite{bertinetto2016fully} & 3.2 & 22.8 & 5.0 & 16.7 & 0.5 & 8.1 & 1.2 & 4.2 & 22.2 & 22.6 & 35.4 & 14.2 & 25.8 & 11.7 & 19.7 & 27.8 & 15.1 & 6.8 & 2.28 & 31.6 & 12.4 & 13.3  \\
SiamRPN~\cite{li2018high} & 1.9 & 15.7 & 4.5 & 12.8 & 1.0 & 1.1 & 6.1 & 8.7 & 7.9 & 6.9 & 17.4 & 17.8 & 20.5 & 7.2 & 18.5 & 5.1 & 9.6 & 15.9 & 15.7 & 21.7 & 3.5 & 14.2  \\
CompNet~\cite{zhang2019comparison} & 28.4 & 41.5 & 65.0 & 66.4 & 37.1 & 49.8 & 16.2 & 31.7 & 69.7 & 73.1 & 75.6 & 71.6 & 61.4 & 52.3 & 63.4 & 39.8 & 52.7 & 75.3 & 60.0 & 47.9 & 25.3 & 52.1 \\
CoAE~\cite{hsieh2019one} & 30.0& 54.9& 64.1& 66.7& 40.1& 54.1& 14.7& 60.9& 77.5& 78.3& 77.9 & 73.2& 80.5& 70.8& \textbf{72.4}& 46.2& 60.1& \textbf{83.9}& 67.1& 75.6& 46.2& 68.2 \\
Li~\etal~\cite{li2020one}  &33.7&58.2&67.5&72.7&40.8&48.2&20.1&55.4&\textbf{78.2}&79.0&76.2&74.6&\textbf{81.3}&71.6 &72.0&48.8& 61.1 &74.3& 68.5 & \textbf{81.0} & 52.4 & 69.1 \\\hline
DETR  &11.4&42.2&44.1&63.4&14.9&40.6&20.6&63.7&62.7&71.5&59.6&52.7&60.6&53.6 &54.9&22.1& 46.2 &62.7& 55.2 & 65.4 & 45.9 & 57.3 \\ 
UP-DETR  &\textbf{46.7}&\textbf{61.2}&\textbf{75.7}&\textbf{81.5}&\textbf{54.8}&\textbf{57.0}&\textbf{44.5}&\textbf{80.7}&74.5&\textbf{86.8}&\textbf{79.1}&\textbf{80.3}&80.6&\textbf{72.0} &70.9&\textbf{57.8}& \textbf{69.0} &80.9& \textbf{71.0} & 80.4 & \textbf{59.9} & \textbf{73.1} \\ \hline\hline
\end{tabular}}
\end{adjustbox}
\end{center}
\end{table*}

\begin{table*}[]
\caption{Panoptic segmentation results on the COCO \texttt{val2017} dataset with the same ResNet-50 backbone. The PanopticFPN++, UPSNet and DETR results are re-implemented by Carion~\etal~\cite{carion2020end}.}
\label{panoptic}
\centering
\begin{tabular}{l|lll|lll|lll|ll}
\hline\hline
Model & PQ & SQ & RQ & PQ$^{th}$ & SQ$^{th}$ & RQ$^{th}$ & PQ$^{st}$ & SQ$^{st}$ & RQ$^{st}$ & AP$^{seg}$ \\\hline
PanopticFPN++~\cite{kirillov2019panoptic} &42.4 & 79.3 &  51.6&49.2 & 82.4 &58.8 &32.3 & 74.8 & 40.6 & \textbf{37.7} \\
UPSNet~\cite{xiong2019upsnet} & 42.5 & 78.0 & 52.5 & 48.6 & 79.4 &59.6 &33.4 &  75.9 &41.7& 34.3 \\
UPSNet-M~\cite{xiong2019upsnet} & 43.0 & 79.1 &  52.8 & 48.9 & 79.7 & 59.7 & 34.1 & 78.2    &42.3 & 34.3 \\\hline
DETR~\cite{carion2020end} & 44.3 & 80.0 & 54.5 & 49.2 & 80.6 &60.3 & 37.0 & 79.1 & 45.9 & 32.9\\
UP-DETR & \textbf{44.7} & \textbf{80.5} & \textbf{54.9} & \textbf{49.7} & \textbf{80.9} & \textbf{60.8} & \textbf{37.2} & \textbf{79.3} & \textbf{46.2} & 34.3 \\ \hline\hline
\end{tabular}
\end{table*}

\figref{fig:cocoAP} shows the AP learning curves on COCO. UP-DETR outperforms DETR for both 150 and 300 epoch schedules with faster convergence. The performance improvement is more noticeable before reducing the learning rate. After reducing the learning rate, UP-DETR still holds the lead of DETR by $\sim$ 1.0 AP improvement. It suggests that the pre-training transformer is still indispensable even on sufficient training data (\ie $\sim$ 118K images on COCO).

\subsection{One-Shot Detection}
\noindent\textbf{Setup.} Given a query image patch whose class label is not included in the training data, one-shot detection aims to detect all instances with the same class in a target image. One-shot detection is a promising research direction that can detect unseen instances. With feeding query patches to the decoder, UP-DETR is naturally compatible to one-shot detection task. Therefore, one-shot detection can also be treated as a downstream fine-tuning task of UP-DETR. 

Following the same one-shot detection setting as~\cite{hsieh2019one}, we crop out the ground truth bounding boxes as the query image patches. 
During fine-tuning, for the given image, we randomly sample a query image patch of a seen class that exists in this image, and the training ground truth is filtered according to the query image patch. 
During evaluation, we first randomly shuffle the query image patches of the class (existing in the image) with a specific random seed of target image ID, then pick up the first five query image patches, and finally average their AP scores. Every existing class in each image is evaluated and then averaged. 
The shuffle procedure ensures the results in stable statistics for evaluation. We train/fine-tune DETR/UP-DETR on VOC \texttt{2007train val} and \texttt{2012train val} sets with 300 epochs, and evaluate on VOC \texttt{2007test} set. We follow the same setting as Li~\etal~\cite{li2020one}.


\noindent\textbf{Results.} ~\tabref{OSD} shows the comparison to the state-of-the-art one-shot detection methods. UP-DETR significantly boosts the performance of DETR on both seen (\textbf{+22.8} AP$^{50}$ gain) and unseen (\textbf{+15.8} AP$^{50}$ gain) classes. Moreover, UP-DETR outperforms all compared methods in both seen (min \textbf{+7.9} AP$^{50}$ gain) and unseen (min \textbf{+4.0} AP$^{50}$ gain) classes of one-shot detection. It further verifies the effectiveness of our pre-training pretext task.

\subsection{Panoptic Segmentation}
The original DETR can be easily extended to panoptic segmentation~\cite{kirillov2019panoptic} by adding a mask head on top of the decoder outputs. Following the same panoptic segmentation training schema as DETR~\cite{carion2020end}, we fine-tune UP-DETR with COCO panoptic segmentation annotation (extra stuff annotation) for only box annotations with 300 epochs. Then, we freeze all the weights of DETR and train the mask head for 25 epochs. We find that UP-DETR also boosts the fine-tuning performance of panoptic segmentation.

\tabref{panoptic} shows the comparison to state-of-the-art methods on panoptic segmentation with the ResNet-50 backbone. As seen, UP-DETR outperforms DETR\footnote{With a bug fixed in github.com/facebookresearch/detr/issues/247, the DETR baseline is better than paper report.} with \textbf{+0.4} PQ, \textbf{+0.5} PQ$^{th}$ and \textbf{+1.4} AP$^{seg}$.

\begin{table}[!t]
\caption{The ablation results of pre-training models with single-query patch and multi-query patches on PASCAL VOC. The values in the brackets are the gaps compared to DETR with the same training schedule.}
\label{number-query-patch}
\centering
\begin{tabular}{l|lll}
\hline\hline
Model & AP   & $\textrm{AP}_{50}$ & $\textrm{AP}_{75}$ \\\hline
DETR & 49.9 & 74.5 & 53.1 \\
UP-DETR (M=1) & 53.1 \small (+3.2) & 77.2 \small (+2.7) & 57.4 \\
UP-DETR (M=10) & \textbf{54.9} \small (+5.0) & \textbf{78.7} \small (+4.2) & \textbf{59.1} \\
\hline\hline
\end{tabular}
\end{table}

\subsection{Ablations}
\label{Analysis}
In ablation experiments, we train/pre-train DETR/UP-DETR models for 15 epochs with the learning rate multiplied by 0.1 at the 10\textit{th} epoch. We fine-tune the UP-DETR models on PASCAL VOC following the setup in \secref{voc} with 150 epochs. Therefore, the average precision results in ablations are relatively lower than those shown in \secref{voc}.

\subsubsection{Single-query patch vs.\ Multi-query patches} 

We pre-train the UP-DETR model with single-query patch ($M=1$) and multi-query patches ($M=10$). Other hyper-parameters are set as mentioned above.

\tabref{number-query-patch} shows the results of single-query patch and multi-query patches. Compared with DETR, UP-DETR surpasses it in all AP metrics by a large margin no matter with single-query patch or multi-query patches. When pre-training UP-DETR with the different number of query patches, UP-DETR ($M=10$) performs better than UP-DETR ($M=1$) on the fine-tuning task, although there are about 2.3 instances per image on PASCAL VOC. Therefore, we adopt the same UP-DETR with $M=10$ for both PASCAL VOC and COCO instead of varying $M$ for different downstream tasks.

\subsubsection{Frozen CNN and Feature Reconstruction} 
\label{freeze_CNN}
To illustrate the importance of frozen CNN backbone and patch feature reconstruction of UP-DETR, we pre-train four different models with different combinations of whether freezing CNN and whether adding feature reconstruction. Note that all the models (including DETR) use the pre-trained CNN on ImageNet.

\begin{table}[!t]
\caption{Ablation study on the frozen CNN and feature reconstruction for pre-training models with $\textrm{AP}_{50}$. The experiments are fine-tuned on PASCAL VOC with 150 epochs.}
\label{Ablation}
\centering
\begin{tabular}{c|c|c|c}
\hline\hline
Case & Frozen CNN & Feature Reconstruction &$\textrm{AP}_{50}$ \\
\hline
DETR & \multicolumn{2}{c|}{scratch transformer} & 74.5 \\\hline
(a) & &  & 74.0 \\
(b) &$\checkmark$ & &  \textbf{78.7} \\
(c)& & $\checkmark$ &  62.0 \\
(d) UP-DETR &$\checkmark$ & $\checkmark$ & \textbf{78.7}\\\hline\hline
\end{tabular}
\end{table}

\begin{figure}[]
\centering
  \includegraphics[width=0.49\textwidth]{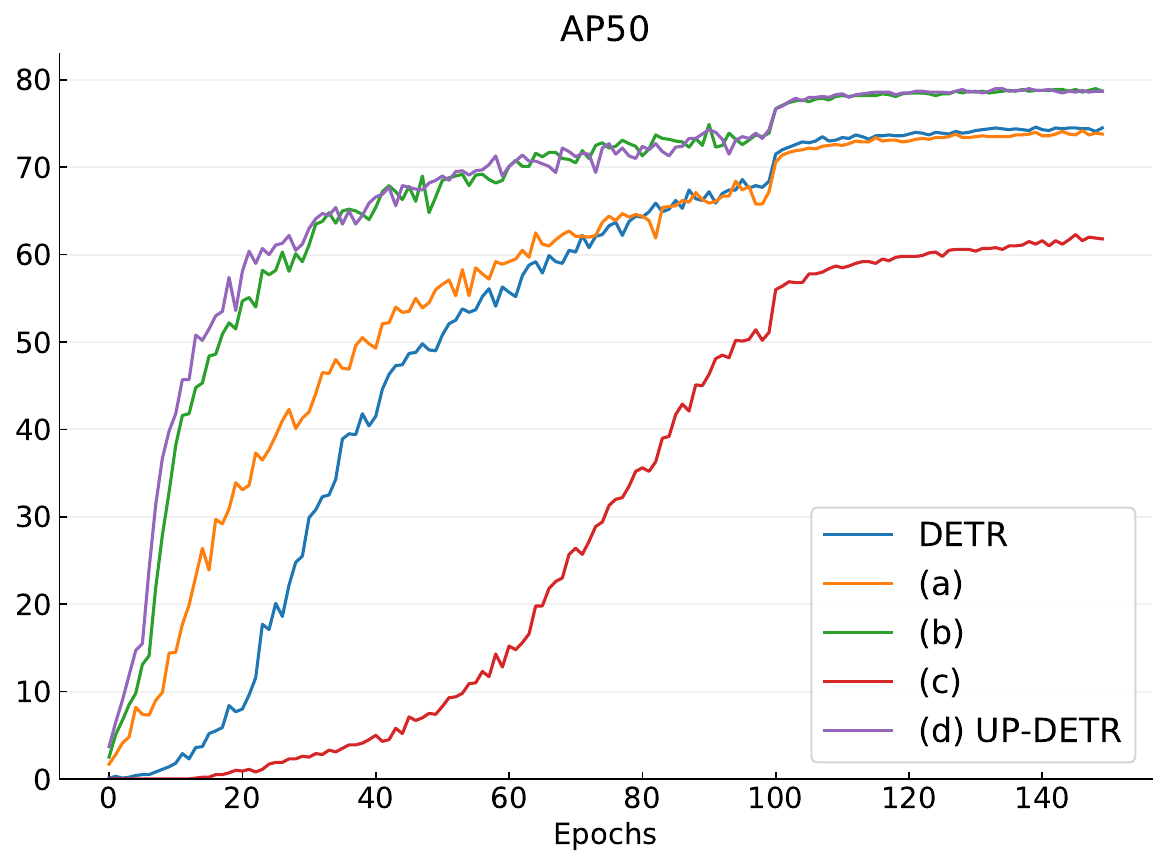}
  \caption{Learning curves ($\textrm{AP}_{50}$) of DETR and four different pre-training models on PASCAL VOC trained with 150 epochs. The models correspond to the models in \tabref{Ablation} one-to-one.}
  \label{5curves}
\end{figure}

\tabref{Ablation} shows $\textrm{AP}_{50}$ of DETR and four different pre-training models on PASCAL VOC with 150 epochs. As shown in \tabref{Ablation}, not all pre-trained models are better than DETR, but pre-training models (b) and (d) perform better than the others. More importantly, without frozen CNN, pre-training models (a) and (c) even perform worse than DETR. This confirms that the frozen pre-training backbone is essential to pre-train the transformer. In addition, it further confirms that the pretext task (random query patch detection) may weaken the feature discrimination of the pre-training CNN, and localization and classification have different feature preferences~\cite{zhang2019towards,wu2020rethinking,song2020revisiting}.

\figref{5curves} plots the $\textrm{AP}_{50}$ learning curves of DETR and four different pre-training models, where the models in \figref{5curves} correspond to the models in \tabref{Ablation} one-to-one. As shown in \figref{5curves}, (d) UP-DETR model achieves faster convergence speed at the early training stage with feature reconstruction. The experiment results suggest that random query patch detection is complementary to the contrastive learning for a better visual representation. The former is designed for the spatial localization with position embeddings, and the latter is designed for instance or cluster classification.

It is worth noting that UP-DETR with frozen CNN and feature reconstruction heavily relies on a pre-trained CNN model,~\eg SwAV CNN. Therefore, we believe that it is a promising direction to further investigate UP-DETR with random query patch detection and contrastive learning together to pre-train the whole DETR model from scratch.

\subsubsection{Attention Mask}
\label{attmask}
After downstream task fine-tuning, we find that there is no noticeable difference between the UP-DETR models pre-trained with and without attention mask. Instead of the fine-tuning results, we plot the loss curves in the pretext task to illustrate the effectiveness of attention mask.

\begin{figure}[!t]
\centering
  \includegraphics[width=0.49\textwidth]{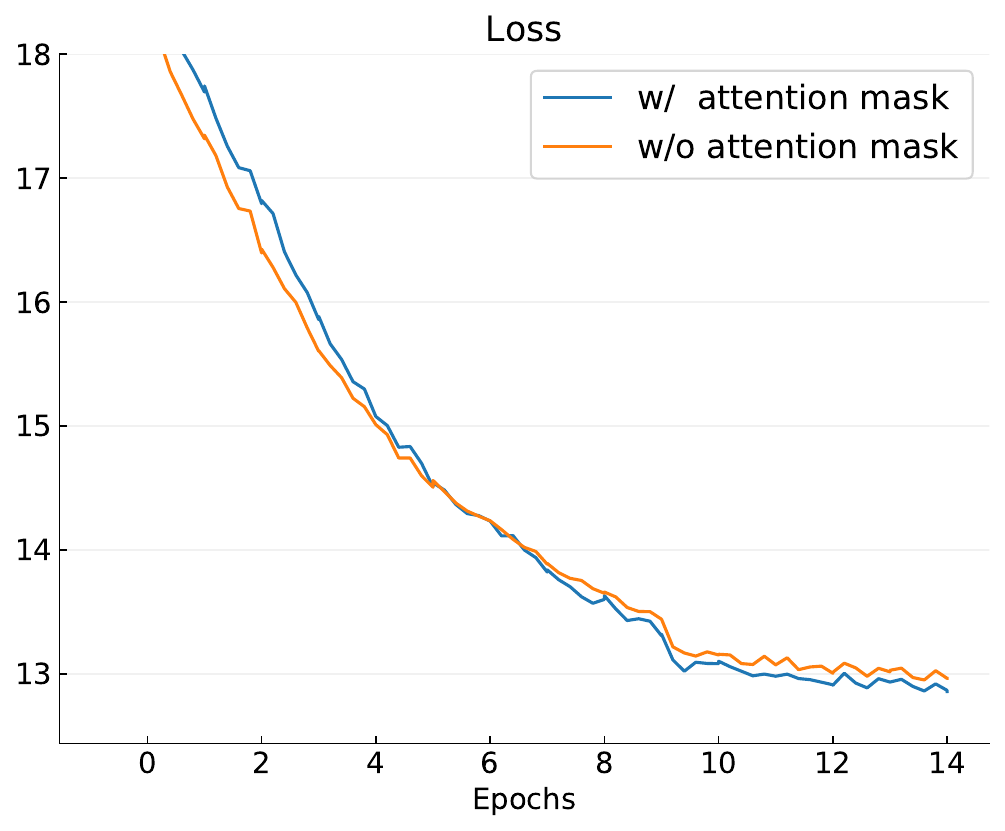}
  \caption{The loss curves of pre-training procedure for UP-DETR w/ and w/o the attention mask.}
  \label{attnmask}
\end{figure}

As shown in \figref{attnmask}, at the early training stage, UP-DETR without attention mask has a lower loss. However, as the model converging, UP-DETR with attention mask overtakes it with a lower loss. The curves seem weird at the first glance, but it is reasonable because the loss is calculated by the optimal bipartite matching. During the early training stage, the model is not converged, and the model without attention mask takes more object queries into attention. Intuitively, the model is easier to be optimized due to introducing more object queries. However, there is a mismatching between the query patch and the ground truth for the model without attention mask. 
As the model converging, the attention mask gradually takes effect, which masks the unrelated query patches and leads to a lower loss.

\begin{figure*}[t]
\centering
  \includegraphics[width=1.0\linewidth]{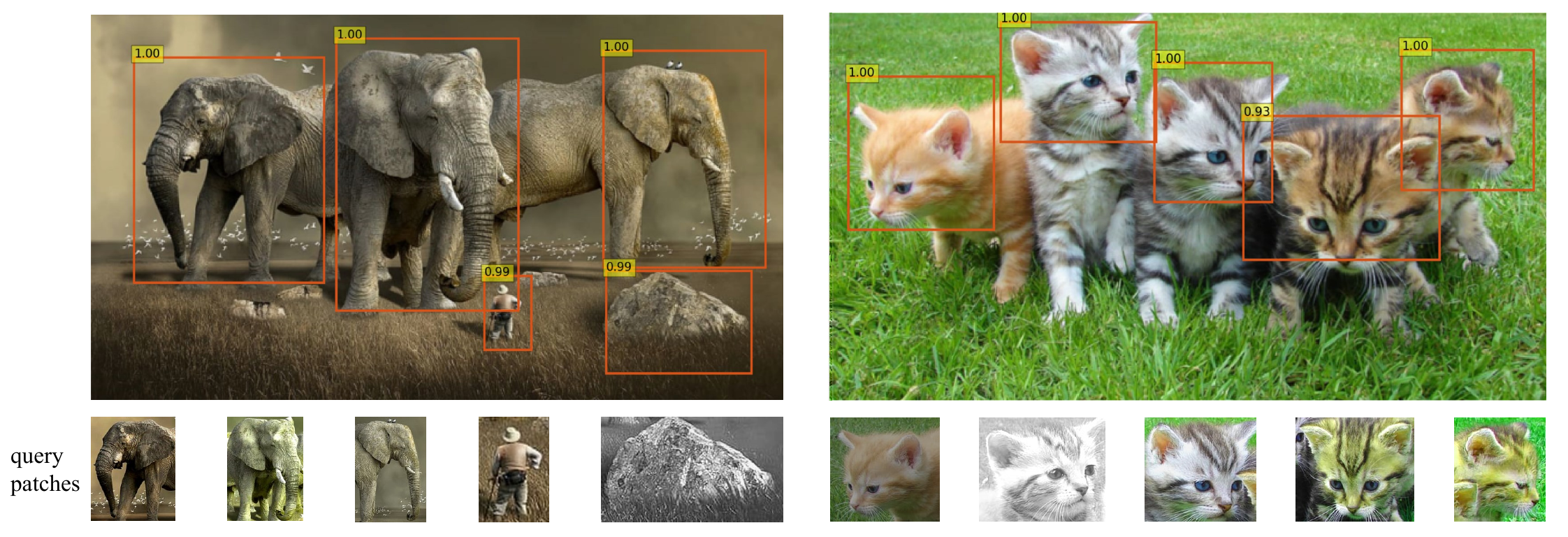}
  \caption{The unsupervised localization of patch queries with UP-DETR. The first line is the original image with predicted bounding boxes. The second line is query patches cropped from the original image with data augmentation (colorjitter and resizing). The value in the upper left corner of the bounding box is the model confidence. As seen, without any annotations, UP-DETR learns to detect patches with given queries in the unsupervised way.}
  \label{vis}
\end{figure*}

\begin{figure*}[t]
\centering
\begin{subfigure}{0.43\linewidth}
\includegraphics[width=1.0\linewidth]{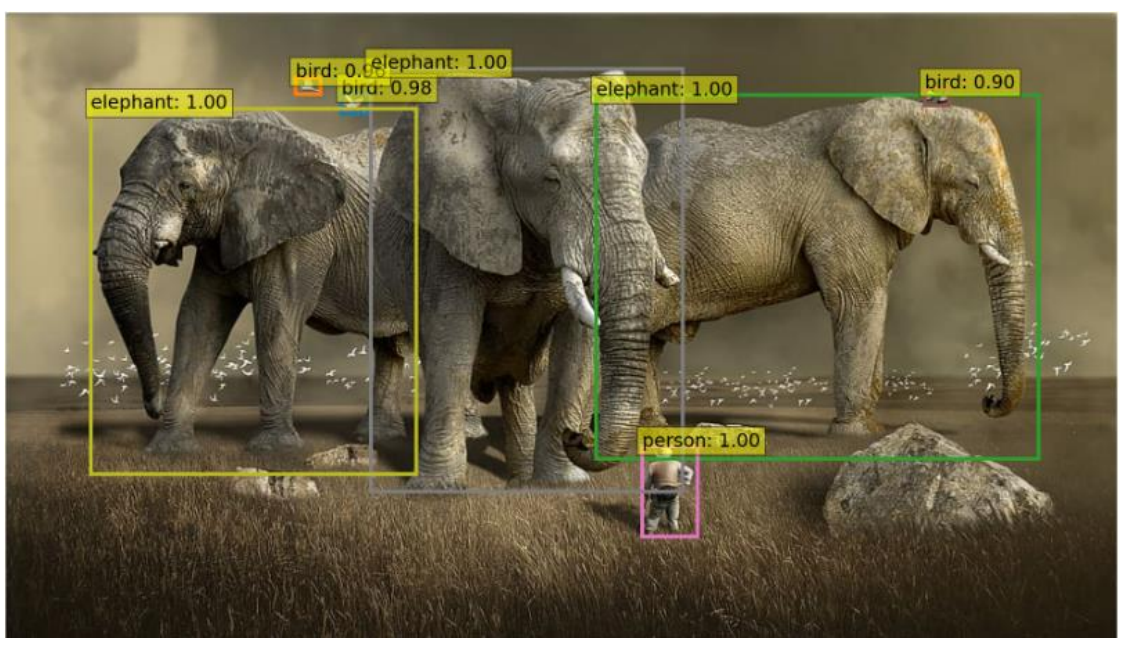} 
\caption{fine-tuned on object detection}
\label{fig:obj-det}
\end{subfigure}
\begin{subfigure}{0.56\linewidth}
\includegraphics[width=1\linewidth]{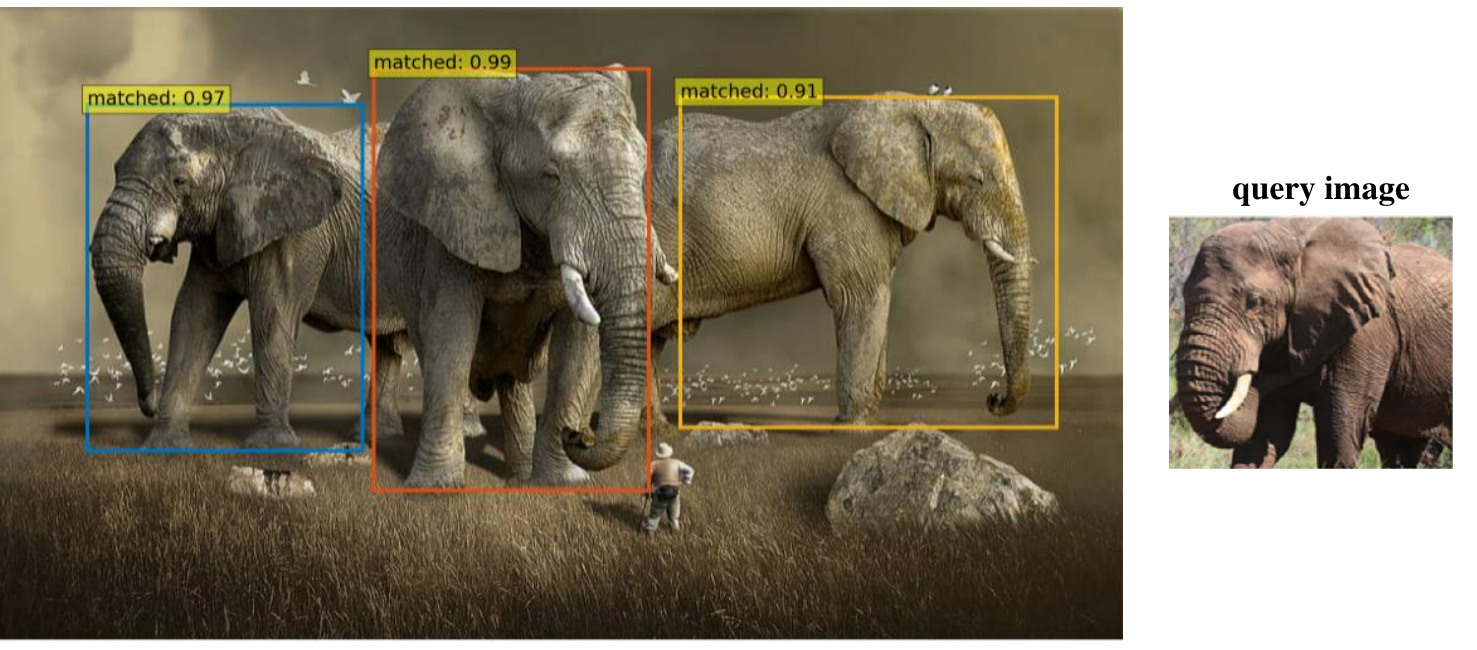}
\caption{fine-tuned on one-shot detection}
\label{fig:one-shot-det}
\end{subfigure}
  \caption{The visualization result of fine-tuned experiments on object detection and one-shot detection. These two models are fine-tuned with the same UP-DETR model. (a) With the input image, the object detection model detects the objects with bounding boxes, whose classes exist in the training set. (b) Given input image and query image, the one-shot detection model detects the objects, which are semantically similar to the query image (with the matched tag instead of the class name). } 
  \label{fine-tune-visual}
\end{figure*}

\subsection{Visualization}
\subsubsection{Pre-training}
To further illustrate the ability of the pre-training model, we visualize the unsupervised localization results of given patch queries. Specifically, for the given image, we manually crop several object patches and apply the data augmentation to them. Then, we feed these patches as queries to the model. Finally, we visualize the model output with bounding boxes, whose classification confidence is greater than $0.9$. This procedure can be treated as \textit{unsupervised one-shot detection} or deep learning based \textit{template matching}.

As shown in \figref{vis}, pre-trained with random query patch detection, UP-DETR successfully learns to locate the bounding box of given query patches and suppresses the duplicated bounding boxes\footnote{Base picture credit: https://www.piqsels.com/en/public-domain-photo-jrkkq, https://www.piqsels.com/en/public-domain-photo-smdfn.}. 
It demonstrates that UP-DETR with random query patch detection is effective to learn the ability of object localization. 

\subsubsection{Fine-tuning}
\figref{fine-tune-visual} shows the visualization result of fine-tuned models on object detection and one-shot detection. Our fine-tuning models perform well on these two tasks. For object detection task, the model is fine-tuned to detect objects with bounding boxes and classes. These classes (\eg elephant, bird, person in \figref{fig:obj-det}) must exist in the training set. Differently, for one-shot detection, given the input image and the query image, the model only predicts the similar objects with bounding boxes and 0/1 labels (matching or not). These two tasks can be unified in our UP-DETR framework by changing the input of the transformer decoder.

\section{Conclusion}
We present a novel pretext task called random query patch detection to unsupervisedly pre-train the transformer in DETR. With unsupervised pre-training, UP-DETR significantly outperforms DETR by a large margin with higher precision and much faster convergence on PASCAL VOC. For the challenging COCO dataset with sufficient training data, UP-DETR still surpasses DETR even with a long training schedule. It indicates that the pre-training transformer is indispensable for different scales of training data in object detection. Besides, UP-DETR also provides a unified perspective for one-shot detection. It significant boosts the performance on one-shot detection task. 

From the perspective of unsupervised pre-training models, the pre-training CNN backbone and the pre-training transformer are separated now. Recent studies of unsupervised pre-training mainly focus on feature discrimination with contrastive learning instead of specialized modules for spatial localization. But in UP-DETR pre-training, the pretext task is mainly designed for patch localization by positional encodings and learnable object queries. We hope an advanced method can integrate the CNN and transformer pre-training into a unified end-to-end framework and apply UP-DETR to more downstream tasks (\eg few-shot object detection and object tracking).

\section*{Acknowledgement}
This work was supported in part by the National Natural Science Foundation of China under Grant 61802130, and in part by the Guangdong Natural Science Foundation under Grant 2019A1515012152 and 2021A1515012651. This work is done when Zhigang Dai was an intern at Tencent Wechat AI.

\ifCLASSOPTIONcaptionsoff
  \newpage
\fi


\bibliographystyle{IEEEtran}
\bibliography{IEEEabrv,egbib}

\vspace{-80pt}
\begin{IEEEbiography}[{\includegraphics[width=1in,height=1.25in,clip,keepaspectratio]{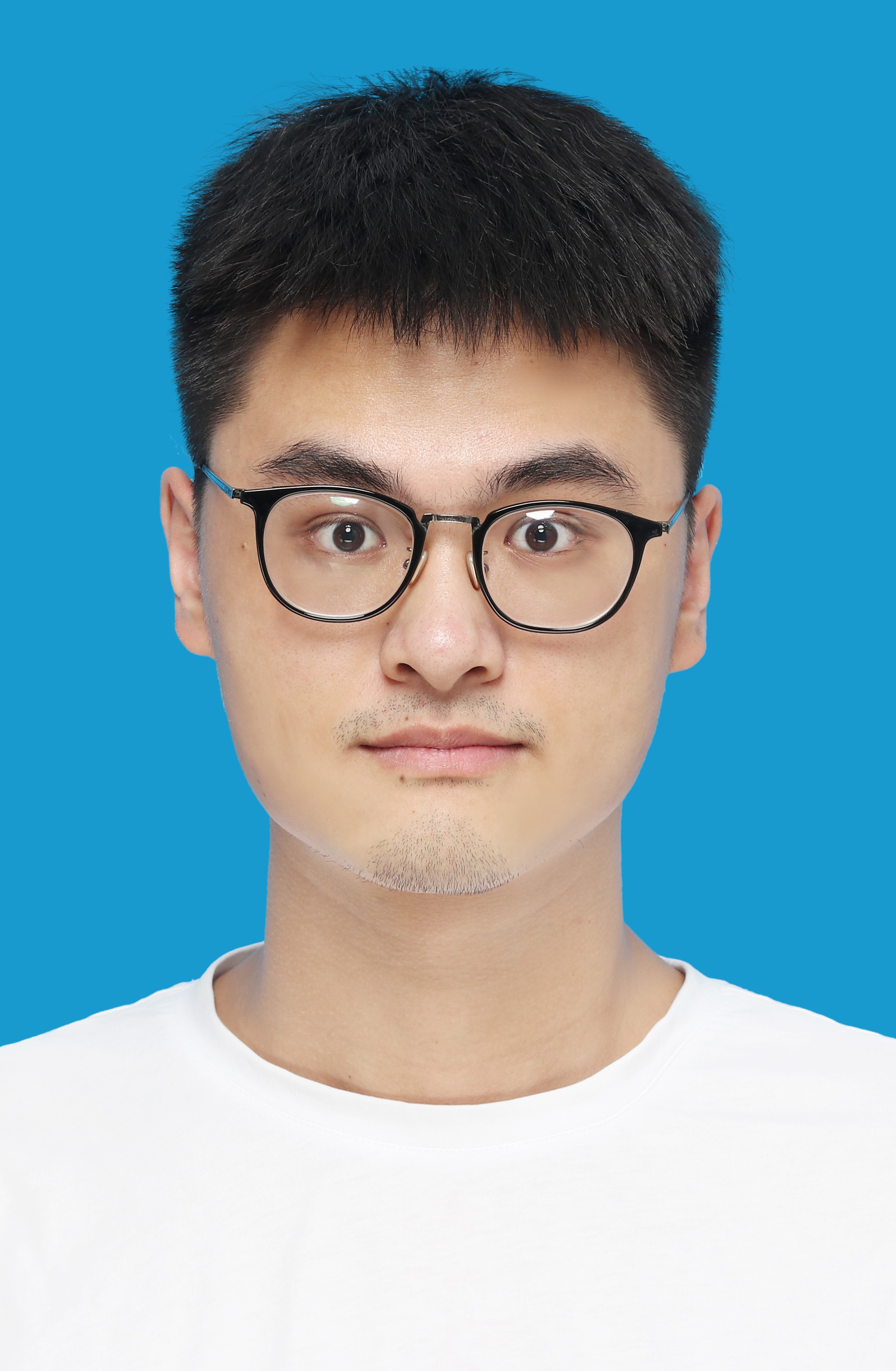}}]{Zhigang Dai} received the B.E. degree from the School of Software Engineering, South China University of Technology, Guangzhou, China, where he is currently pursuing the master's degree.

His research interests include visual representation learning and object detection. He used to be an intern at Tencent Wechat AI and published a paper at CVPR 2021 as an oral presentation.
\end{IEEEbiography}

\vspace{30pt}
\begin{IEEEbiography}[{\includegraphics[width=1in,height=1.25in,clip,keepaspectratio]{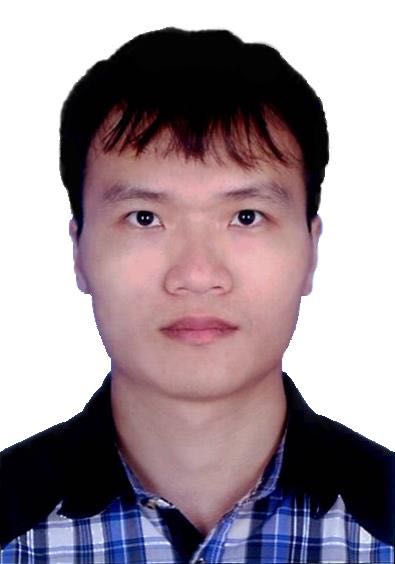}}]{Bolun Cai} received the M.Eng. and Ph.D. degrees from the South China University of Technology, Guangzhou, China, in 2016 and 2019, respectively. He is currently a Senior Researcher with Tencent WeChat AI.

His research interests include computer vision, machine learning, and image processing.
\end{IEEEbiography}

\vspace{-330pt}
\begin{IEEEbiography}[{\includegraphics[width=1in,height=1.25in,clip,keepaspectratio]{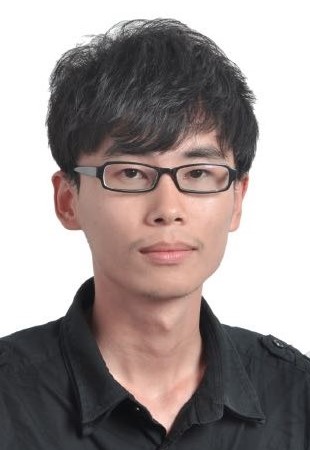}}]{Yugeng Lin} received the B.Eng. degree from the Sun Yat-sen University in 2012. He is currently an Expert Researcher with Tencent WeChat AI.

His research interests include computer vision and image processing.
\end{IEEEbiography}

\vspace{-330pt}
\begin{IEEEbiography}[{\includegraphics[width=1in,height=1.25in,clip,keepaspectratio]{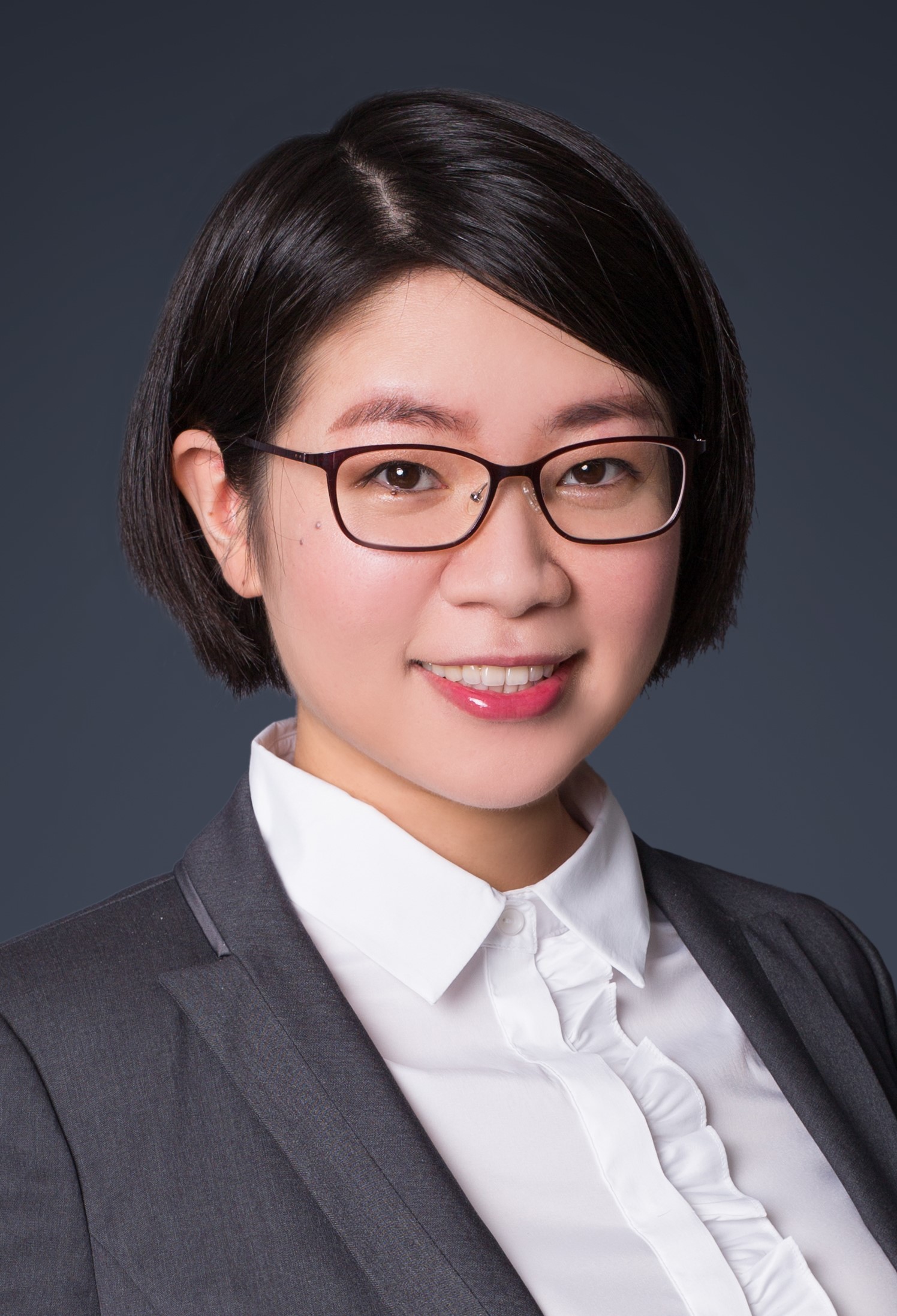}}]{Junying Chen} (Senior Member, IEEE) received the B.E. degree in electronic and information engineering from Zhejiang University, Hangzhou, China, in 2007, and the Ph.D. degree in electrical and electronic engineering from The University of Hong Kong, Hong Kong, in 2013. She is currently an Associate Professor with the School of Software Engineering, South China University of Technology, Guangzhou, China. She is the member of IEEE-HKN, and the senior members of IEEE, CCF, CSIG, and CSBME.

She published over 40 papers in academic journals and conferences, including IEEE TNNLS, IEEE TMI, IEEE JBHI, CVPR, ACM MM, ACL, etc. She received the Second Prize for Technological Invention from CCF Science and Technology Award, and the Second Prize of Excellent Papers of Guangdong Computer Academy. Her research interests include deep learning, neural networks, computer vision, and image processing.
\end{IEEEbiography}

\end{document}